\documentclass[10pt]{article}
\PassOptionsToPackage{table}{xcolor}
\usepackage[preprint]{tmlr}

\usepackage{amsmath,amssymb,amsthm,mathtools}
\usepackage{graphicx}
\usepackage{enumitem}
\usepackage{hyperref}
\usepackage{cleveref}
\usepackage{booktabs}
\usepackage{xcolor}
\usepackage{float}
\usepackage{caption}
\usepackage{pgfplots}
\pgfplotsset{compat=1.18}

\newtheorem{theorem}{Theorem}[section]

\theoremstyle{definition}

\theoremstyle{remark}
\newtheorem{remark}[theorem]{Remark}

\newcommand{\R}{\mathbb{R}}
\newcommand{\Ds}{\mathcal{D}_s}
\newcommand{\Dc}{\mathcal{D}_c}
\newcommand{\rank}{\operatorname{rank}}

\title{Measuring Alignment-Induced Activation Shifts Correctly:\\
  A Template-Controlled Difference-in-Differences Protocol}

\author{\name Yuki Nakamura \\
        \addr The Open University of Japan \\
        \addr ORCID: \texttt{\href{https://orcid.org/0009-0001-7174-6737}{0009-0001-7174-6737}}}

\begin{document}
\maketitle

\begin{abstract}
Comparing a model's internal activations before and after alignment is
a natural way to ask what safety training changes: one forms the matrix
of paired aligned-minus-base activations on safety-relevant inputs and
reads off its effective rank or top direction.  We show the obvious way
to form this matrix is confounded.  The aligned model is evaluated
under a chat template the base model never saw, so the naive
difference conflates the alignment shift with chat formatting.  We
introduce a four-variant decomposition of the modification matrix---%
naive, template-controlled, within-aligned, and
difference-in-differences (DiD)---that separates the two effects.
Template control alone removes a $2.0$--$3.9\times$ inflation of the
measured effective rank across Llama-3.1-8B, Gemma-2-9B, and
Qwen-2.5-7B; the DiD contrast is what recovers the refusal direction
of \citet{arditi2024refusal}, lifting its cosine alignment from
$0.18$--$0.39$ to $0.50$--$0.86$.  Projection-ablation across the
three families confirms the recovered subspace is behaviorally active
and that singular-value order is not causal order.  We validate the
protocol on a controlled testbed and distill it into measurement
recommendations for activation-difference studies of alignment.
\end{abstract}

\section{Introduction}
\label{sec:intro}

Comparing a model's internal representations before and after
alignment is a natural lens on what safety fine-tuning changes.  A
large body of work localizes safety-relevant behavior through
differences of activations: \citet{arditi2024refusal} take the
difference of the aligned model's mean activations on harmful versus
harmless prompts to extract a single ``refusal direction'' whose
ablation removes refusal, and representation-engineering
methods~\citep{zou2023representation, turner2023steering} steer on
within-model contrast vectors.  A complementary and equally natural
object is the alignment modification matrix itself: the matrix of
paired $(\text{aligned}-\text{base})$ activations on a chosen input
distribution, whose effective rank and principal structure summarize
how concentrated the alignment update is.  This is the object we
study.

We show that the obvious way to form this matrix is confounded, and we
give the fix.  The confound is mundane but consequential.  The
aligned (instruction-tuned) model is evaluated under a chat template;
the base model has never seen that template and is evaluated on raw
text.  The naive difference therefore measures
\emph{alignment plus chat formatting}, not alignment.  The formatting
component is large: it inflates the measured effective rank by
$2.0$--$3.9\times$ across three model families
(\S\ref{sec:families}), and it changes which direction a
top-singular-vector analysis recovers.

\paragraph{The protocol.}
We decompose the modification matrix into four variants
(\S\ref{sec:protocol}) collected under matched conditions:
$M_{\mathrm{naive}}$ (the default: aligned-under-template minus
base-under-raw), $M_{\mathrm{template}}$ (both checkpoints under the
same template), $M_{\mathrm{aligned}}$ (the within-aligned template
shift), and $M_{\mathrm{DiD}}$ (a difference-in-differences contrast
that further subtracts the control-input shift).  The variants answer
two different questions, and conflating them is the source of the
confound:
\begin{itemize}[nosep]
  \item \textbf{Template control corrects the \emph{rank}.}  Moving
        from $M_{\mathrm{naive}}$ to $M_{\mathrm{template}}$ removes
        the formatting variance and lowers the effective-rank ratio
        $\rho_{\epsilon}$ by $2.0$--$3.9\times$ (Llama $3.9$, Gemma
        $2.0$, Qwen $2.2$; \S\ref{sec:families}).
  \item \textbf{The DiD contrast recovers the \emph{direction}.}
        Template control alone does \emph{not} recover the Arditi
        refusal direction (cosine stays $0.18$--$0.39$); the DiD
        contrast does, lifting the cosine to $0.50$ (Qwen), $0.77$
        (Llama), and $0.86$ (Gemma).
\end{itemize}
A projection-ablation test across all three families
(\S\ref{sec:causal}) confirms the recovered low-rank subspace is
behaviorally load-bearing---ablating it collapses refusal while a
random subspace of the same rank does not---and, as a caution for the
practice of reading structure off the spectrum, that the
singular-value ordering is not the causal ordering.

\paragraph{Contributions.}
(1) We identify and quantify a chat-template confound in
activation-difference measurements of alignment, and give a
four-variant decomposition that isolates formatting from alignment and
separates the rank question from the direction question
(\S\ref{sec:protocol}).  (2) We validate on a controlled testbed that
the corrected effective rank tracks a planted rank and is a diagnostic
of the shift's structure rather than a quantity that can be inflated
on demand (\S\ref{sec:validation}).  (3) We apply the protocol to
three open-weight families and report the corrected rank, the
direction recovery, and a causal validation, surfacing
family-dependent structure the naive protocol hides
(\S\ref{sec:families}--\ref{sec:causal}).  (4) We distill the result
into concrete measurement recommendations
(\S\ref{sec:conclusion}).  Everything is computable from paired
checkpoints by SVD alone, with no training; code and data are
released (\S\ref{sec:availability}).

\section{The Modification Matrix and What Measuring the Shift Requires}
\label{sec:matrix}

Let $f_{\mathrm{pre}}, f_{\mathrm{align}}$ be the pre- and
post-alignment systems, and $h_{\mathrm{pre}}(x),
h_{\mathrm{align}}(x) \in \R^d$ their residual-stream activations at a
fixed layer for input $x$.  Given a sample $x_1,\dots,x_n$ from an
input distribution $\mathcal{D}$, the \emph{alignment modification
matrix} is
\begin{equation}
  M_{\mathcal{D}}
  := \bigl[\,
       h_{\mathrm{align}}(x_1) - h_{\mathrm{pre}}(x_1),
       \;\ldots,\;
       h_{\mathrm{align}}(x_n) - h_{\mathrm{pre}}(x_n)
     \,\bigr]
  \;\in\; \R^{d\times n}.
  \label{eq:modification}
\end{equation}
With singular values $\sigma_1 \geq \sigma_2 \geq \cdots \geq 0$, the
\emph{effective rank} at tolerance $\epsilon \in (0,1)$ is the number
of components needed to capture a $(1-\epsilon)$ fraction of the
modification's variance,
\begin{equation}
  \rank_{\epsilon}(M_{\mathcal{D}})
  := \min\Bigl\{\, k :
       \textstyle\sum_{i=1}^{k}\sigma_i^2
       \,\geq\, (1-\epsilon)\textstyle\sum_i \sigma_i^2 \Bigr\},
  \qquad
  \rho_{\epsilon}(M_{\mathcal{D}})
  := \frac{\rank_{\epsilon}(M_{\mathcal{D}})}{d}.
  \label{eq:effrank}
\end{equation}
We use $\epsilon = 0.05$ throughout.  A refusal-direction analysis is
the $k{=}1$ special case (the top left singular vector $u_1$); an
effective-rank readout is $\rho_{\epsilon}$.  Both are functions of
$M_{\mathcal{D}}$, so both inherit whatever confound enters
$M_{\mathcal{D}}$.

\paragraph{The chat-template confound.}
The instruction-tuned checkpoint $f_{\mathrm{align}}$ is normally
evaluated with its chat template applied (system/user turn markers,
special tokens); the base checkpoint $f_{\mathrm{pre}}$ ships without
one and is evaluated on raw text.  When one then differences the two,
\[
  M_{\mathrm{naive}} =
    h_{\mathrm{align}}(x_{\mathrm{chat}}) - h_{\mathrm{pre}}(x_{\mathrm{raw}}),
\]
the result mixes the effect of alignment with the effect of the
formatting tokens the aligned model was conditioned on and the base
model was not.  This is the quantity one obtains when a base model is
differenced against its instruct sibling without matching the input
formatting.  It is not a measurement of the alignment shift;
\S\ref{sec:protocol} separates the two components, and
\S\ref{sec:families} shows the formatting component dominates the
effective-rank estimate.

\section{The Four-Variant Protocol}
\label{sec:protocol}

To separate alignment from formatting we collect activations for each
of $\{\text{base},\text{aligned}\}\times\{\text{raw},\text{chat-template}\}$
on the same prompts (the base tokenizer is given the aligned model's
template so both are formatted identically), and form four
modification matrices:
\begin{align*}
  M_{\mathrm{naive}}    &= h_{\mathrm{align}}(x_{\mathrm{chat}}) - h_{\mathrm{pre}}(x_{\mathrm{raw}})
     &&\text{(the default; alignment {\bf+} formatting),}\\
  M_{\mathrm{template}} &= h_{\mathrm{align}}(x_{\mathrm{chat}}) - h_{\mathrm{pre}}(x_{\mathrm{chat}})
     &&\text{(formatting matched on both checkpoints),}\\
  M_{\mathrm{aligned}}  &= h_{\mathrm{align}}(x_{\mathrm{chat}}) - h_{\mathrm{align}}(x_{\mathrm{raw}})
     &&\text{(the within-aligned template shift, for reference),}\\
  M_{\mathrm{DiD}}      &= M_{\mathrm{template}}(\Ds) - \overline{M_{\mathrm{template}}(\Dc)}
     &&\text{(control-subtracted difference-in-differences),}
\end{align*}
where $\Ds$ is a safety-relevant input distribution, $\Dc$ a matched
benign control, and $\overline{(\cdot)}$ the column-mean over $\Dc$.

The protocol's value is that the four variants answer two distinct
questions, and the literature's default ($M_{\mathrm{naive}}$)
conflates them.  \textbf{(i) How concentrated is the alignment shift?}
This is a question about \emph{rank}, and $M_{\mathrm{template}}$ is
the right quantity: matching the template removes the formatting
variance that otherwise inflates $\rho_{\epsilon}$.  \textbf{(ii)
Which direction does alignment write along?}  This is a question about
the \emph{top direction}, and we show (\S\ref{sec:families}) that
$M_{\mathrm{template}}$ is \emph{not} enough---its top singular vector
is still dominated by a residual formatting mean shared by safety and
control inputs.  Subtracting the control shift ($M_{\mathrm{DiD}}$) is
what isolates the safety-specific direction and recovers the Arditi
refusal direction.  We make this split explicit because it is easy to
assume that one correction does both jobs; it does not.

\begin{remark}[What $\rho_{\epsilon}$ does and does not certify]
\label{rem:scope}
$\rho_{\epsilon}$ is a readout-level measurement of how concentrated
the alignment modification is at the residual stream; it is silent on
the depth of the computation that produces the
shift~\citep{olah2020zoom, wollschlager2025}, and low rank is a
generic property of linearly represented concepts, not a
safety-specific signature (\S\ref{sec:consequence}).  The bridge from
a recovered low-rank subspace to behavior is the causal ablation of
\S\ref{sec:causal}, not the rank number itself.
\end{remark}

\section{Validating the Protocol on a Controlled Testbed}
\label{sec:validation}

Before applying the protocol to LLMs, we check on a controlled testbed
that the corrected effective rank measures what it claims to---the
structural concentration of the shift---and is not an artifact of the
estimator or a quantity that can be trivially inflated.

We train a 3-layer MLP ($d_{\mathrm{hidden}}{=}128$, input dimension
$20$, $N_{\mathrm{train}}{=}4000$) on a binary task
$y_{\mathrm{task}}=\mathbf{1}[X_4X_5 - X_6X_7 > 0]$ with
$X\sim\mathcal{N}(0,I_{20})$, and impose a ``safety'' constraint on a
disjoint corner $\Ds=\{X_1,X_2,X_3>1\}$ where the safe label is forced
to $0$ (safety and task features do not overlap).  We then align the
network by three procedures of increasing modification rank:
\textbf{Steering} (a rank-$1$ activation overlay gated to $\Ds$),
\textbf{Full fine-tune}, and \textbf{Distributed} (full fine-tune plus
a rank-maximization regularizer: a stable-rank surrogate
$\|M\|_F^2/\|M\|_2^2$ with column-orthogonality, strength $\lambda$;
App.~\ref{app:calibration}).  Measured on $\Ds$, $\rho_{\epsilon}$
moves monotonically with the procedure, from $0.008$ (Steering) to
$0.17$ (Full FT) to $0.33$--$0.40$ (Distributed): the corrected
measurement tracks the planted concentration.

\paragraph{$\rho_{\epsilon}$ is a diagnostic, not a dial.}
A natural worry is that one could simply optimize $\rho_{\epsilon}$
upward.  The testbed shows this does not buy structural robustness.
Sweeping the regularizer strength $\lambda\in\{0.5,5,15,50\}$
(Table~\ref{tab:lambda_sweep}, App.~\ref{app:calibration}), the same
nominal $\rho_{\epsilon}\approx0.40$ admits two qualitatively
different alignments: at $\lambda{=}5$ compliance is robust to
projecting out $62.5\%$ of the representation ($1.00$ at $r{=}80$,
$d{=}128$), while at $\lambda{=}50$, at the same nominal rank,
compliance collapses ($0.10$ at $r{=}40$).  Mechanically inflating the
rank past the point where it tracks the safety signal buys neither
robustness nor compliance.  $\rho_{\epsilon}$ thus diagnoses the
shift's structure but is not itself a training target---which is why
the right use of the protocol is measurement and audit, not
optimization.

\begin{table}[h]
\centering
\caption{Controlled testbed: the same nominal
$\rho_{\epsilon}\approx0.40$ in the Distributed family is
ablation-robust at $\lambda{=}5$ and brittle at $\lambda{=}50$.
Compliance on $\Ds$ after projecting out the top-$r$ principal
directions of $M_{\Ds}$; MLP $d{=}128$, mean $\pm$ std over $3$ seeds;
baseline is pre-ablation compliance.  Source:
\texttt{calibration\_v2\_revision3.json}.}
\label{tab:lambda_sweep}
\begin{tabular}{@{}lcccc@{}}
\toprule
$\lambda$ & $\rho_{\epsilon}$ & Baseline
  & Compliance at $r{=}40$ & Compliance at $r{=}80$ \\
\midrule
$0.5$ & $0.33{\scriptstyle\pm 0.00}$ & $0.94{\scriptstyle\pm 0.01}$
      & $1.00{\scriptstyle\pm 0.00}$ & $1.00{\scriptstyle\pm 0.00}$ \\
$5$   & $0.40{\scriptstyle\pm 0.01}$ & $0.86{\scriptstyle\pm 0.03}$
      & $\mathbf{1.00}{\scriptstyle\pm 0.00}$ & $\mathbf{1.00}{\scriptstyle\pm 0.00}$ \\
$15$  & $0.40{\scriptstyle\pm 0.01}$ & $0.78{\scriptstyle\pm 0.02}$
      & $1.00{\scriptstyle\pm 0.00}$ & $0.68{\scriptstyle\pm 0.30}$ \\
$50$  & $0.40{\scriptstyle\pm 0.01}$ & $0.71{\scriptstyle\pm 0.02}$
      & $\mathbf{0.10}{\scriptstyle\pm 0.14}$ & $0.14{\scriptstyle\pm 0.07}$ \\
\bottomrule
\end{tabular}
\end{table}

\section{Applying the Protocol to Three LLM Families}
\label{sec:families}

We apply the protocol to Llama-3.1-8B vs.\ Llama-3.1-8B-Instruct
($d{=}4096$, $L{=}32$), Gemma-2-9B vs.\ Gemma-2-9B-it ($d{=}3584$,
$L{=}42$), and Qwen-2.5-7B vs.\ Qwen-2.5-7B-Instruct ($d{=}3584$,
$L{=}28$).  Activations are bf16, residual stream, last prompt token,
$n{=}200$ safety-relevant ($\Ds$) and $n{=}200$ control ($\Dc$)
inputs.  $\Ds$ is a curated set of AdvBench-style harmful-behavior
requests spanning cybercrime, weapons, fraud, harassment, illegal
substances, privacy, deception, and manipulation; $\Dc$ is a matched
set of benign factual queries (prompt lists in
\texttt{experiments/prompts\_v4.py}).

\subsection{The confound is large, and family-dependent}
\label{sec:families_rank}

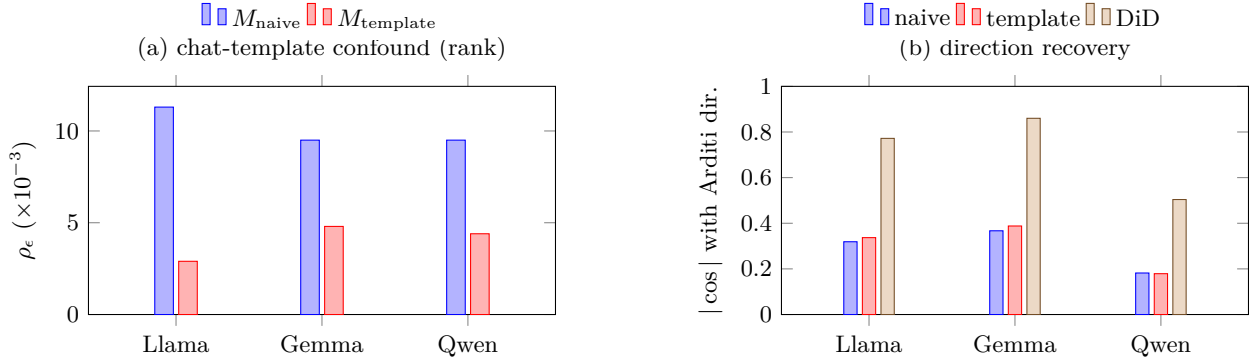
\begin{figure}[t]
\centering
\begin{tikzpicture}
\begin{axis}[
  width=0.47\textwidth, height=4.6cm,
  ybar, bar width=7pt,
  symbolic x coords={Llama,Gemma,Qwen}, xtick=data,
  ymin=0, ylabel={$\rho_{\epsilon}$ ($\times 10^{-3}$)},
  title={(a) chat-template confound (rank)},
  title style={font=\small}, label style={font=\small}, tick label style={font=\small},
  legend style={at={(0.5,1.20)},anchor=south,legend columns=-1,font=\small,draw=none},
  enlarge x limits=0.3,
]
\addplot coordinates {(Llama,11.3) (Gemma,9.5) (Qwen,9.5)};
\addplot coordinates {(Llama,2.9) (Gemma,4.8) (Qwen,4.4)};
\legend{$M_{\mathrm{naive}}$, $M_{\mathrm{template}}$}
\end{axis}
\end{tikzpicture}\hfill
\begin{tikzpicture}
\begin{axis}[
  width=0.47\textwidth, height=4.6cm,
  ybar, bar width=5pt,
  symbolic x coords={Llama,Gemma,Qwen}, xtick=data,
  ymin=0, ymax=1, ylabel={$|\cos|$ with Arditi dir.},
  title={(b) direction recovery},
  title style={font=\small}, label style={font=\small}, tick label style={font=\small},
  legend style={at={(0.5,1.20)},anchor=south,legend columns=-1,font=\small,draw=none},
  enlarge x limits=0.3,
]
\addplot coordinates {(Llama,0.319) (Gemma,0.367) (Qwen,0.182)};
\addplot coordinates {(Llama,0.337) (Gemma,0.388) (Qwen,0.179)};
\addplot coordinates {(Llama,0.772) (Gemma,0.860) (Qwen,0.504)};
\legend{naive, template, DiD}
\end{axis}
\end{tikzpicture}
\caption{The confound and its correction across three families.
\textbf{(a)} Template control lowers the effective-rank ratio
$\rho_{\epsilon}$ by $2.0$--$3.9\times$ (the chat-template confound).
\textbf{(b)} Template control leaves the Arditi refusal direction
unrecovered ($|\cos|$ essentially unchanged from naive); the DiD
contrast recovers it.  Values from
Tables~\ref{tab:four_variants} and~\ref{tab:arditi}.}
\label{fig:confound}
\end{figure}

Table~\ref{tab:four_variants} reports $\rho_{\epsilon}$ for all four
variants on $\Ds$.  Moving from the naive measurement to the
template-controlled one lowers the effective-rank ratio by
$3.9\times$ (Llama), $2.0\times$ (Gemma), and $2.2\times$ (Qwen): a
large fraction of what a naive analysis attributes to alignment is
chat formatting.  The magnitude is itself family-dependent---largest
on Llama (Fig.~\ref{fig:confound}a)---which is structure the naive
protocol hides under a single inflated number.

\begin{table}[h]
\centering
\caption{Four-variant $\rho_{\epsilon}$ on $\Ds$ ($\epsilon{=}0.05$,
mean across hidden layers, $n{=}200$).  The naive$\to$template drop is
the chat-template confound; $M_{\mathrm{aligned}}$ (within-aligned
template shift) is reported for reference.  Source:
\texttt{v4\_\{llama,gemma,qwen\}.json}.}
\label{tab:four_variants}
\begin{tabular}{@{}lcccccc@{}}
\toprule
Model & $d$ & $M_{\mathrm{naive}}$ & $M_{\mathrm{template}}$
  & $M_{\mathrm{aligned}}$ & $M_{\mathrm{DiD}}$
  & naive/template \\
\midrule
Llama-3.1-8B-Instruct & $4096$ & $0.0113$ & $0.0029$ & $0.0133$ & $0.0101$ & $3.9\times$ \\
Gemma-2-9B-it         & $3584$ & $0.0095$ & $0.0048$ & $0.0168$ & $0.0119$ & $2.0\times$ \\
Qwen-2.5-7B-Instruct  & $3584$ & $0.0095$ & $0.0044$ & $0.0112$ & $0.0166$ & $2.2\times$ \\
\bottomrule
\end{tabular}
\end{table}

The corrected ($M_{\mathrm{template}}$) effective rank is small in
absolute terms: $12$, $17$, $16$ components out of $d\geq 3584$ capture
$95\%$ of the variance (Table~\ref{tab:rank}).  But it is not as small
as a single direction.  Subtracting the column mean before SVD raises
the residual effective rank to $\sim76$--$80$
(Table~\ref{tab:rank}, last column), $4$--$7\times$ the uncentered
value, so a single shared mean direction captures most of the measured
concentration, with a multi-dimensional remainder consistent with the
concept-cone picture of \citet{wollschlager2025}.  This is why
$M_{\mathrm{template}}$ corrects the rank but not the direction: its
top singular vector is dominated by that mean, which is shared between
$\Ds$ and $\Dc$ and is therefore not safety-specific.

\begin{table}[h]
\centering
\caption{Corrected ($M_{\mathrm{template}}$) effective rank on $\Ds$
at $\epsilon{=}0.05$, mean across hidden layers, $n{=}200$.  The
centered column subtracts the column mean before SVD; its
$4$--$7\times$ larger value localizes the bulk of the measured rank to
a single mean direction.}
\label{tab:rank}
\begin{tabular}{@{}lccccc@{}}
\toprule
Model & $d$
  & $\overline{\rank_{\epsilon}}(\Ds)$ & $\rho_{\epsilon}$
  & $\rho_{\epsilon}(\Dc)$ & centered $\overline{\rank_{\epsilon}}$ \\
\midrule
Llama-3.1-8B-Instruct & $4096$ & $12.0$ & $0.0029$ & $0.0059$ & $77.6$ \\
Gemma-2-9B-it         & $3584$ & $17.4$ & $0.0048$ & $0.0096$ & $76.2$ \\
Qwen-2.5-7B-Instruct  & $3584$ & $15.7$ & $0.0044$ & $0.0050$ & $80.4$ \\
\bottomrule
\end{tabular}
\end{table}

\subsection{The DiD contrast recovers the refusal direction; template
control alone does not}
\label{sec:families_direction}

\citet{arditi2024refusal} define a refusal direction as the
within-aligned difference of means
$\bar{\Delta}^{\mathrm{Arditi}} =
\mathbb{E}_{\Ds}[h_{\mathrm{align}}] -
\mathbb{E}_{\Dc}[h_{\mathrm{align}}]$.
Table~\ref{tab:arditi} and Figure~\ref{fig:confound}b report the
absolute cosine between this direction and the top singular vector of
each variant.  Neither the
naive nor the template-controlled matrix recovers it (cosine
$0.18$--$0.39$): both top vectors are dominated by the shared mean
shift.  The DiD contrast, which subtracts that shared shift, recovers
it---to $0.77$ (Llama) and $0.86$ (Gemma), and more moderately to
$0.50$ (Qwen, max $0.80$ across layers).  The recovery is therefore a
property of the \emph{contrast}, not of template control; and the
single-direction picture is family-dependent, strong on Llama and
Gemma but only moderate on Qwen.

\begin{table}[h]
\centering
\caption{Absolute cosine between the Arditi refusal direction and the
top singular vector of each variant (mean over hidden layers).
Template control corrects the rank but leaves the direction
unrecovered; the DiD contrast recovers it.  Source:
\texttt{v4\_\{llama,gemma,qwen\}.json}.}
\label{tab:arditi}
\begin{tabular}{@{}lccc@{}}
\toprule
Model & vs $M_{\mathrm{naive}}$ & vs $M_{\mathrm{template}}$
      & vs $M_{\mathrm{DiD}}$ \\
\midrule
Llama-3.1-8B-Instruct & $0.319$ & $0.337$ & $\mathbf{0.772}$ \\
Gemma-2-9B-it         & $0.367$ & $0.388$ & $\mathbf{0.860}$ \\
Qwen-2.5-7B-Instruct  & $0.182$ & $0.179$ & $\mathbf{0.504}$ \\
\bottomrule
\end{tabular}
\end{table}

\section{Causal Validation: SVD Order Is Not Causal Order}
\label{sec:causal}

A recovered low-rank subspace is only useful if it is behaviorally
load-bearing.  We test this directly.  For each family we project out
the top-$k$ principal subspace $P_k = U_kU_k^\top$ of
$M_{\mathrm{template}}$ at the residual stream (substituting
$h_{\mathrm{align}}\to(I-P_k)h_{\mathrm{align}}$ across a layer band)
and measure refusal on $n_{\mathrm{gen}}{=}100$ held-out harmful
prompts, against a random rank-$k$ control at the same layers.
Refusal is scored by a $15$-keyword case-insensitive substring match
(full list in the released code; an LLM-judge cross-check on Llama
agrees at $100\%$, App.~\ref{app:judge}).

\begin{table}[h]
\centering
\caption{Causal projection-ablation at the narrow band
$[0.45L,0.70L]$, $k{=}3$, $n_{\mathrm{gen}}{=}100$, Wilson $95\%$ CIs.
Ablating the recovered subspace collapses refusal where a random
subspace of the same rank does not.}
\label{tab:ablation}
\begin{tabular}{@{}lccc@{}}
\toprule
Model & Baseline & Principal rank-$3$ & Random rank-$3$ \\
\midrule
Llama-3.1-8B-Instruct & $0.97$ & $0.80$ \;[$0.71,0.87$] & $0.94\pm0.00$ \\
Gemma-2-9B-it         & $0.99$ & $\mathbf{0.00}$ \;[$0.00,0.04$] & $0.99\pm0.00$ \\
Qwen-2.5-7B-Instruct  & $0.96$ & $0.54$ \;[$0.44,0.63$] & $0.95$ \;[$0.89,0.98$] \\
\bottomrule
\end{tabular}
\end{table}

The recovered subspace is causally privileged in every family: a
principal rank-$k$ projection drives refusal into a Wilson $95\%$ CI
$\subset[0.00,0.06]$ (Gemma at narrow $k{=}1$; Qwen at narrow
$k{\geq}20$, monotone $0.92\to0.54\to0.24\to0.06\to0.01$; Llama at
wide-band $k{=}20$), while a random subspace of the same rank leaves
refusal intact (Table~\ref{tab:ablation},
App.~\ref{app:ablation_sweep}).  The rank at which collapse occurs is
family-dependent---one direction for Gemma, ${\sim}10$ for Qwen, a
wider-band block for Llama---mirroring the direction-recovery
heterogeneity of \S\ref{sec:families_direction}.

\paragraph{The spectral order is not the causal order.}
A caution follows for the common practice of reading importance off
the singular-value spectrum.  On Llama, per-direction ablation
(Table~\ref{tab:perdir}) shows $u_1$ alone drops refusal to $0.84$ but
$u_2$ alone has near-zero effect \emph{despite ranking second by
$\sigma$}, and the cumulative narrow-band curve is non-monotone at
$k{=}5$ (refusal rebounds to $0.92$ before collapsing at wider $k$;
App.~\ref{app:ablation_sweep}).  Variance order and causal order
coincide for Gemma but not for Llama.  The implication for the
protocol is concrete: the behavioral relevance of a recovered subspace
must be measured by ablation as a curve in $k$, not inferred from
where a direction sits in the spectrum.

\begin{table}[h]
\centering
\caption{Per-direction narrow-band ablation on Llama (baseline
$0.97$).  Each row ablates a single direction $u_i$; the $u_3$--$u_5$
row reports the common $0.94$ obtained for each individually.}
\label{tab:perdir}
\begin{tabular}{@{}lcc@{}}
\toprule
Ablate & Refusal & Note \\
\midrule
$u_1$ alone & $0.84$ & refusal-promoting \\
$u_2$ alone & $0.99$ & near-zero causal effect \\
each of $u_3$--$u_5$ alone & $0.94$ & weak \\
\bottomrule
\end{tabular}
\end{table}

\section{What the Corrected Measurement Implies}
\label{sec:consequence}

Two scope points keep the interpretation honest.  First, low rank is
\emph{generic}: on the same models, arbitrary linear concepts
(English-vs-French, question-vs-statement) give
$\rho_{\epsilon}\approx0.008$--$0.010$, only $1.9$--$2.7\times$ the
safety value (App.~\ref{app:lrh}).  A small effective rank is thus a
property of linearly represented concepts in
general~\citep{park2024linear, marks2023geometry}, and is not by
itself a safety signature; the safety-specific content is supplied by
the causal ablation of \S\ref{sec:causal}, not by the rank number.
Second, the measurement is at the readout, not the mechanism
(Remark~\ref{rem:scope}): a distributed non-linear computation writing
into a low-dimensional residual subspace tests as low-rank.

Within that scope, a corrected low $\rho_{\epsilon}$ carries one
direct structural consequence, which is exactly what makes the
measurement worth getting right.

\begin{remark}[Linear-projection fragility]
\label{rem:fragility}
If the corrected modification is $(k,\epsilon)$-concentrated on $\Ds$
(its top-$k$ principal subspace captures $(1-\epsilon)$ of the
variance), an adversary with whitebox activation access and the base
checkpoint can remove $(1-\epsilon)$ of the alignment shift with a
rank-$k$ projection $P_k=U_kU_k^\top$, replacing
$h_{\mathrm{align}}(x)$ with $(I-P_k)h_{\mathrm{align}}(x)+P_k
h_{\mathrm{pre}}(x)$.  The estimate $\hat U_k$ is a truncated SVD on a
finite sample ($n{=}200$ suffices on $d\geq3584$;
App.~\ref{app:bootstrap}).  A \emph{naive} measurement, inflating $k$
by formatting variance, would over-state the dimensionality of this
attack surface by $2$--$4\times$; measuring it correctly matters for
reasoning about the attack at all.
\end{remark}

This is a statement about residual-stream projection, the operational
attack that representation-engineering
methods~\citep{zou2023representation, turner2023steering} already
perform; it does not speak to LoRA fine-tuning, prompt jailbreaks, or
behavioral deception, which are out of scope.

\section{Related Work}
\label{sec:related}

\paragraph{Refusal directions and activation differencing.}
\citet{arditi2024refusal} introduced the within-aligned
difference-of-means refusal direction and showed ablating it removes
refusal; \citet{wollschlager2025} showed refusal is mediated by
concept \emph{cones}---multiple mechanistically independent
directions, not a single one---consistent with the multi-dimensional
residual we recover after centering (\S\ref{sec:families_rank}).
Representation engineering~\citep{zou2023representation,
turner2023steering} steers and ablates along such difference vectors;
\citet{jain2024what, lee2024mechanistic} give mechanistic accounts of
safety fine-tuning and DPO~\citep{rafailov2023dpo}.  Our contribution
is orthogonal and methodological: these analyses difference activations
\emph{within} a single model to localize behaviors, whereas our
aligned-vs-base modification matrix is a complementary object; we show
its obvious naive form is confounded by chat formatting and give a
protocol that separates the rank question from the direction question.  Sparse
autoencoders~\citep{templeton2024scaling} offer an alternative
decomposition basis; whether SAE features yield a more causally
aligned subspace than SVD components on $M_{\mathrm{template}}$ is a
natural question (we use SVD because it is computable from paired
checkpoints alone).

\paragraph{Fragility of safety fine-tuning.}
Safety alignment is empirically
fragile~\citep{qi2023finetuning, lermen2023lora, qi2024safety,
betley2025emergent} and elastically reverts under
pressure~\citep{ji2025elasticity}.  These behavioral findings share a
geometric prerequisite---a low-rank, projectable modification
(Remark~\ref{rem:fragility})---which the corrected measurement is the
right tool to quantify.

\paragraph{Linear representations and concept erasure.}
LLMs represent many concepts
linearly~\citep{mikolov2013linguistic, park2024linear,
marks2023geometry, hernandez2024linearity}; our low corrected
$\rho_{\epsilon}$ is consistent with this, and the LRH baseline
(App.~\ref{app:lrh}) shows alignment is only modestly more
concentrated than generic concepts.  Concept
erasure~\citep{ravfogel2020null, belrose2023leace} bounds
linear-projection removal; weight-space low-rank
adaptation~\citep{hu2022lora} and mode
connectivity~\citep{garipov2018loss, frankle2020linear} explain why
small parameter perturbations~\citep{lermen2023lora} can undo
alignment.  Broader alignment context:
\citet{ouyang2022, bai2022, christiano2017} on RLHF/preference
training, \citet{ngo2024alignment} on the deep-learning view of
alignment, and \citet{greenblatt2024alignment, hubinger2024sleeper} on
alignment faking, whose representational study we leave to future work.

\section{Conclusion: Measurement Recommendations}
\label{sec:conclusion}

Activation-difference analyses of alignment are only as trustworthy as
the matrix they start from.  We have shown the default matrix is
confounded by chat formatting---inflating the measured effective rank
by $2.0$--$3.9\times$ and leaving the refusal direction unrecovered---%
and that a four-variant decomposition separates formatting from
alignment and the rank question from the direction question.  We
distill the protocol into three recommendations for studies that
difference aligned and base activations:
\begin{enumerate}[nosep]
  \item \textbf{Match the formatting.}  Apply the aligned model's chat
        template to \emph{both} checkpoints ($M_{\mathrm{template}}$)
        before reading off an effective rank; the naive difference
        over-states concentration by $2$--$4\times$.
  \item \textbf{Use a difference-in-differences contrast for the
        direction.}  Template control corrects the rank but not the
        direction; subtract the control-input shift
        ($M_{\mathrm{DiD}}$) to recover a safety-specific direction.
  \item \textbf{Validate causally, per family.}  Confirm a recovered
        subspace by projection-ablation against a random control, as a
        curve in $k$; do not read causal importance off the
        singular-value order, and report per-family rather than
        assuming a universal single-direction structure.
\end{enumerate}
The protocol is computable by SVD from paired checkpoints with no
training (the capability burden is in constructing $\Ds$, not in the
measurement).  The most useful extension is empirical reach---more
model families and alignment procedures (SFT/RLHF/DPO and beyond)
measured under the corrected protocol.

\section*{Code and Data Availability}
\label{sec:availability}

All experiment code, prompt sets, and pre-computed JSON results used in
this paper are publicly available at
\url{https://github.com/Nakammura/effective-rank-audit} under the
CC-BY-4.0 license.  The repository contains the LaTeX source of this
paper; the diagnostic pipeline (\texttt{diagnostics\_v4.py},
\texttt{qwen\_four\_variant.py}); the calibration MLP
(\texttt{calibration\_v2.py}, \texttt{calibration\_v2\_revision3.py});
Modal wrappers for cloud-GPU reproduction (\texttt{modal\_qwen.py},
\texttt{modal\_revision3.py}); and the LLM-judge cross-check
(\texttt{llm\_judge\_eval.py}).  An immutable snapshot of the v1.0
release is archived on Zenodo:
\href{https://doi.org/10.5281/zenodo.20341445}{\texttt{10.5281/zenodo.20341445}}
(concept DOI for all versions:
\href{https://doi.org/10.5281/zenodo.20341444}{\texttt{10.5281/zenodo.20341444}}).

\bibliography{references}

\begin{thebibliography}{29}
\providecommand{\natexlab}[1]{#1}
\providecommand{\url}[1]{\texttt{#1}}
\expandafter\ifx\csname urlstyle\endcsname\relax
  \providecommand{\doi}[1]{doi: #1}\else
  \providecommand{\doi}{doi: \begingroup \urlstyle{rm}\Url}\fi

\bibitem[Arditi et~al.(2024)Arditi, Obeso, Syed, Paleka, Panickssery, Gurnee,
  and Nanda]{arditi2024refusal}
Andy Arditi, Oscar Obeso, Aaquib Syed, Daniel Paleka, Nina Panickssery, Wes
  Gurnee, and Neel Nanda.
\newblock Refusal in language models is mediated by a single direction.
\newblock In \emph{Advances in Neural Information Processing Systems},
  volume~37, 2024.

\bibitem[Bai et~al.(2022)Bai, Kadavath, Kundu, Askell, Kernion,
  et~al.]{bai2022}
Yuntao Bai, Saurav Kadavath, Sandipan Kundu, Amanda Askell, Jackson Kernion,
  et~al.
\newblock Constitutional {AI}: Harmlessness from {AI} feedback.
\newblock 2022.
\newblock arXiv preprint arXiv:2212.08073.

\bibitem[Belrose et~al.(2023)Belrose, Schneider-Joseph, Ravfogel, Cotterell,
  Raff, and Biderman]{belrose2023leace}
Nora Belrose, David Schneider-Joseph, Shauli Ravfogel, Ryan Cotterell, Edward
  Raff, and Stella Biderman.
\newblock {LEACE}: Perfect linear concept erasure in closed form.
\newblock In \emph{Advances in Neural Information Processing Systems}, 2023.
\newblock arXiv preprint arXiv:2306.03819.

\bibitem[Betley et~al.(2026)Betley, Warncke, Sztyber-Betley, Tan, Bao, Soto,
  Srivastava, Labenz, and Evans]{betley2025emergent}
Jan Betley, Niels Warncke, Anna Sztyber-Betley, Daniel Tan, Xuchan Bao, Martin
  Soto, Megha Srivastava, Nathan Labenz, and Owain Evans.
\newblock Training large language models on narrow tasks can lead to broad
  misalignment.
\newblock \emph{Nature}, 649:\penalty0 584--589, 2026.

\bibitem[Christiano et~al.(2017)Christiano, Leike, Brown, Martic, Legg, and
  Amodei]{christiano2017}
Paul~F. Christiano, Jan Leike, Tom Brown, Miljan Martic, Shane Legg, and Dario
  Amodei.
\newblock Deep reinforcement learning from human preferences.
\newblock \emph{Advances in Neural Information Processing Systems}, 30, 2017.

\bibitem[Frankle et~al.(2020)Frankle, Dziugaite, Roy, and
  Carbin]{frankle2020linear}
Jonathan Frankle, Gintare~Karolina Dziugaite, Daniel~M. Roy, and Michael
  Carbin.
\newblock Linear mode connectivity and the lottery ticket hypothesis.
\newblock In \emph{International Conference on Machine Learning}, 2020.
\newblock arXiv preprint arXiv:1912.05671.

\bibitem[Garipov et~al.(2018)Garipov, Izmailov, Podoprikhin, Vetrov, and
  Wilson]{garipov2018loss}
Timur Garipov, Pavel Izmailov, Dmitrii Podoprikhin, Dmitry~P. Vetrov, and
  Andrew~Gordon Wilson.
\newblock Loss surfaces, mode connectivity, and fast ensembling of {DNN}s.
\newblock In \emph{Advances in Neural Information Processing Systems}, 2018.
\newblock arXiv preprint arXiv:1802.10026.

\bibitem[Greenblatt et~al.(2024)Greenblatt, Denison, Wright, Roger,
  et~al.]{greenblatt2024alignment}
Ryan Greenblatt, Carson Denison, Benjamin Wright, Fabien Roger, et~al.
\newblock Alignment faking in large language models.
\newblock 2024.
\newblock arXiv preprint arXiv:2412.14093.

\bibitem[Hernandez et~al.(2024)Hernandez, Sen~Sharma, Haklay, Meng, Wattenberg,
  Andreas, Belinkov, and Bau]{hernandez2024linearity}
Evan Hernandez, Arnab Sen~Sharma, Tal Haklay, Kevin Meng, Martin Wattenberg,
  Jacob Andreas, Yonatan Belinkov, and David Bau.
\newblock Linearity of relation decoding in transformer language models.
\newblock In \emph{International Conference on Learning Representations}, 2024.
\newblock arXiv preprint arXiv:2308.09124.

\bibitem[Hu et~al.(2022)Hu, Shen, Wallis, Allen-Zhu, Li, Wang, Wang, and
  Chen]{hu2022lora}
Edward~J. Hu, Yelong Shen, Phillip Wallis, Zeyuan Allen-Zhu, Yuanzhi Li, Shean
  Wang, Lu~Wang, and Weizhu Chen.
\newblock Lo{RA}: Low-rank adaptation of large language models.
\newblock In \emph{International Conference on Learning Representations}, 2022.
\newblock arXiv preprint arXiv:2106.09685.

\bibitem[Hubinger et~al.(2024)Hubinger, Denison, Mu, Lambert, Tong,
  et~al.]{hubinger2024sleeper}
Evan Hubinger, Carson Denison, Jesse Mu, Mike Lambert, Meg Tong, et~al.
\newblock Sleeper agents: Training deceptive {LLMs} that persist through safety
  training.
\newblock 2024.
\newblock arXiv preprint arXiv:2401.05566.

\bibitem[Jain et~al.(2024)Jain, Lubana, Oksuz, Joy, Torr, Sanyal, and
  Dokania]{jain2024what}
Samyak Jain, Ekdeep~Singh Lubana, Kemal Oksuz, Tom Joy, Philip~H.S. Torr,
  Amartya Sanyal, and Puneet~K. Dokania.
\newblock What makes and breaks safety fine-tuning? {A} mechanistic study.
\newblock In \emph{Advances in Neural Information Processing Systems},
  volume~37, 2024.

\bibitem[Ji et~al.(2025)Ji, Wang, Qiu, Chen, Zhou, Li, Lou, Dai, Liu, and
  Yang]{ji2025elasticity}
Jiaming Ji, Kaile Wang, Tianyi Qiu, Boyuan Chen, Jiayi Zhou, Changye Li, Hantao
  Lou, Juntao Dai, Yunhuai Liu, and Yaodong Yang.
\newblock Language models resist alignment: Evidence from data compression.
\newblock In \emph{Proceedings of the 63rd Annual Meeting of the Association
  for Computational Linguistics}, 2025.
\newblock Best Paper Award.

\bibitem[Lee et~al.(2024)Lee, Bai, Pres, Wattenberg, Kummerfeld, and
  Mihalcea]{lee2024mechanistic}
Andrew Lee, Xiaoyan Bai, Itamar Pres, Martin Wattenberg, Jonathan~K.
  Kummerfeld, and Rada Mihalcea.
\newblock A mechanistic understanding of alignment algorithms: A case study on
  {DPO} and toxicity.
\newblock In \emph{International Conference on Machine Learning}, 2024.

\bibitem[Lermen et~al.(2023)Lermen, Rogers-Smith, and Ladish]{lermen2023lora}
Simon Lermen, Charlie Rogers-Smith, and Jeffrey Ladish.
\newblock {LoRA} fine-tuning efficiently undoes safety training in {L}lama
  2-chat 70{B}.
\newblock 2023.
\newblock arXiv preprint arXiv:2310.20624.

\bibitem[Marks \& Tegmark(2024)Marks and Tegmark]{marks2023geometry}
Samuel Marks and Max Tegmark.
\newblock The geometry of truth: Emergent linear structure in large language
  model representations of true/false datasets.
\newblock In \emph{Conference on Language Modeling (COLM)}, 2024.

\bibitem[Mikolov et~al.(2013)Mikolov, Yih, and Zweig]{mikolov2013linguistic}
Tomas Mikolov, Wen-tau Yih, and Geoffrey Zweig.
\newblock Linguistic regularities in continuous space word representations.
\newblock In \emph{Proceedings of NAACL-HLT}, 2013.

\bibitem[Ngo et~al.(2024)Ngo, Chan, and Mindermann]{ngo2024alignment}
Richard Ngo, Lawrence Chan, and S\"{o}ren Mindermann.
\newblock The alignment problem from a deep learning perspective.
\newblock In \emph{International Conference on Learning Representations}, 2024.

\bibitem[Olah et~al.(2020)Olah, Cammarata, Schubert, Goh, Petrov, and
  Carter]{olah2020zoom}
Chris Olah, Nick Cammarata, Ludwig Schubert, Gabriel Goh, Michael Petrov, and
  Shan Carter.
\newblock Zoom in: An introduction to circuits.
\newblock \emph{Distill}, 2020.
\newblock \doi{10.23915/distill.00024.001}.

\bibitem[Ouyang et~al.(2022)Ouyang, Wu, Jiang, Almeida, Wainwright,
  et~al.]{ouyang2022}
Long Ouyang, Jeffrey Wu, Xu~Jiang, Diogo Almeida, Carroll Wainwright, et~al.
\newblock Training language models to follow instructions with human feedback.
\newblock \emph{Advances in Neural Information Processing Systems}, 35, 2022.

\bibitem[Park et~al.(2024)Park, Choe, and Veitch]{park2024linear}
Kiho Park, Yo~Joong Choe, and Victor Veitch.
\newblock The linear representation hypothesis and the geometry of large
  language models.
\newblock In \emph{International Conference on Machine Learning}, 2024.
\newblock arXiv preprint arXiv:2311.03658.

\bibitem[Qi et~al.(2024)Qi, Zeng, Xie, Chen, Jia, Mittal, and
  Henderson]{qi2023finetuning}
Xiangyu Qi, Yi~Zeng, Tinghao Xie, Pin-Yu Chen, Ruoxi Jia, Prateek Mittal, and
  Peter Henderson.
\newblock Fine-tuning aligned language models compromises safety, even when
  users do not intend to.
\newblock In \emph{International Conference on Learning Representations}, 2024.

\bibitem[Qi et~al.(2025)Qi, Panda, Lyu, Ma, Roy, Beirami, Mittal, and
  Henderson]{qi2024safety}
Xiangyu Qi, Ashwinee Panda, Kaifeng Lyu, Xiao Ma, Subhrajit Roy, Ahmad Beirami,
  Prateek Mittal, and Peter Henderson.
\newblock Safety alignment should be made more than just a few tokens deep.
\newblock In \emph{International Conference on Learning Representations}, 2025.
\newblock Outstanding Paper Award.

\bibitem[Rafailov et~al.(2023)Rafailov, Sharma, Mitchell, Ermon, Manning, and
  Finn]{rafailov2023dpo}
Rafael Rafailov, Archit Sharma, Eric Mitchell, Stefano Ermon, Christopher~D.
  Manning, and Chelsea Finn.
\newblock Direct preference optimization: Your language model is secretly a
  reward model.
\newblock \emph{Advances in Neural Information Processing Systems}, 36, 2023.

\bibitem[Ravfogel et~al.(2020)Ravfogel, Elazar, Gonen, Twiton, and
  Goldberg]{ravfogel2020null}
Shauli Ravfogel, Yanai Elazar, Hila Gonen, Michael Twiton, and Yoav Goldberg.
\newblock Null it out: Guarding protected attributes by iterative nullspace
  projection.
\newblock In \emph{Annual Meeting of the ACL}, 2020.

\bibitem[Templeton et~al.(2024)Templeton, Conerly, Marcus,
  et~al.]{templeton2024scaling}
Adly Templeton, Tom Conerly, Jonathan Marcus, et~al.
\newblock Scaling monosemanticity: Extracting interpretable features from
  {Claude 3 Sonnet}.
\newblock Transformer Circuits Thread, 2024.
\newblock \url{https://transformer-circuits.pub/2024/scaling-monosemanticity/}.

\bibitem[Turner et~al.(2023)Turner, Thiergart, Leech, Udell, Vazquez, Mini, and
  MacDiarmid]{turner2023steering}
Alexander~Matt Turner, Lisa Thiergart, Gavin Leech, David Udell, Juan~J.
  Vazquez, Ulisse Mini, and Monte MacDiarmid.
\newblock Steering language models with activation engineering.
\newblock 2023.
\newblock arXiv preprint arXiv:2308.10248.

\bibitem[Wollschl\"{a}ger et~al.(2025)Wollschl\"{a}ger, Elstner, Geisler,
  Cohen-Addad, G\"{u}nnemann, and Gasteiger]{wollschlager2025}
Tom Wollschl\"{a}ger, Jannes Elstner, Simon Geisler, Vincent Cohen-Addad,
  Stephan G\"{u}nnemann, and Johannes Gasteiger.
\newblock The geometry of refusal in large language models: Concept cones and
  representational independence.
\newblock In \emph{International Conference on Machine Learning}, 2025.

\bibitem[Zou et~al.(2023)Zou, Phan, Chen, Campbell, Guo,
  et~al.]{zou2023representation}
Andy Zou, Long Phan, Sarah Chen, James Campbell, Phillip Guo, et~al.
\newblock Representation engineering: A top-down approach to {AI} transparency.
\newblock 2023.
\newblock arXiv preprint arXiv:2310.01405.

\end{thebibliography}
\bibliographystyle{tmlr}

\appendix

\section{Effective rank: bootstrap stability and sample-size sweep}
\label{app:bootstrap}

We bootstrap-resample the $n{=}200$ paired safety samples with
replacement ($200$ resamples per layer) and recompute
$\rank_{\epsilon}$ on $M_{\mathrm{template}}$ at $\epsilon{=}0.05$.
The resampling distribution is tight ($\pm1$ to $\pm2$ across all
layers and families) and centered at or below the full-sample
(without-replacement) point estimate---the standard downward bias of
with-replacement bootstrap on rank functionals (resampling reduces the
effective unique sample to ${\approx}0.63n$).  The full-sample column
is the rank reported in the body (Table~\ref{tab:bootstrap_ci}).

\begin{table}[h]\centering
\caption{Bootstrap-resample distribution (with replacement,
$n_{\mathrm{boot}}{=}200$) of $\rank_{\epsilon}(M_{\mathrm{template}})$
at $\epsilon{=}0.05$, vs.\ the full-sample point estimate at $n{=}200$.
Source: \texttt{v4\_llama.json}, \texttt{v4\_gemma.json},
\texttt{revision6\_qwen\_bootstrap.json}.}
\label{tab:bootstrap_ci}
\begin{tabular}{@{}llcccc@{}}
\toprule
Family & Layer & Full-sample & Boot.\ mean & Boot.\ std & Boot.\ 95\% range \\
\midrule
Llama (L=32) & 14 & $19$ & $15.42$ & $0.66$ & $[14,17]$ \\
             & 16 & $17$ & $14.51$ & $0.62$ & $[13,16]$ \\
             & 18 & $13$ & $11.47$ & $0.58$ & $[11,12]$ \\
\midrule
Gemma (L=42) & 19 & $27$ & $21.68$ & $0.92$ & $[20,23]$ \\
             & 21 & $24$ & $19.40$ & $0.87$ & $[18,21]$ \\
             & 23 & $23$ & $18.56$ & $0.82$ & $[17,20]$ \\
\midrule
Qwen (L=28)  & 13 & $~5$ & $~5.04$ & $0.23$ & $[5,~6]$ \\
             & 18 & $19$ & $15.67$ & $0.61$ & $[15,17]$ \\
             & 24 & $49$ & $35.91$ & $1.22$ & $[34,38]$ \\
\bottomrule
\end{tabular}
\end{table}

\paragraph{Sample-size sweep.}
Sweeping $n\in\{50,100,200\}$ on $M_{\mathrm{template}}$ at a
representative middle layer ($L{=}16$ Llama, $L{=}21$ Gemma), the
effective rank saturates well below $\min(d,n)$
(Table~\ref{tab:n_sweep}): the rank is signal-driven, not sample-bound.

\begin{table}[h]\centering
\caption{Sample-size sweep, $\epsilon{=}0.05$.
$\rank_{\epsilon}/\min(d,n)$ stays $<0.25$ at $n{=}200$.}
\label{tab:n_sweep}
\begin{tabular}{@{}lccc@{}}
\toprule
Model, $n$ & $\rank_{\epsilon}$ & $\rho_{\epsilon}$ & $\rank_{\epsilon}/\min(d,n)$ \\
\midrule
Llama, $n{=}50$  & $9$  & $0.0022$ & $0.180$ \\
Llama, $n{=}100$ & $16$ & $0.0039$ & $0.160$ \\
Llama, $n{=}200$ & $17$ & $0.0042$ & $0.085$ \\
\midrule
Gemma, $n{=}50$  & $12$ & $0.0033$ & $0.240$ \\
Gemma, $n{=}100$ & $22$ & $0.0061$ & $0.220$ \\
Gemma, $n{=}200$ & $24$ & $0.0067$ & $0.120$ \\
\bottomrule
\end{tabular}
\end{table}

\paragraph{$\epsilon$-sensitivity.}
The headline uses $\epsilon{=}0.05$.  The narrow-band mean effective
rank decreases monotonically with $\epsilon$; at $\epsilon{=}0.20$
Llama and Gemma collapse to rank-$1$ (the mean direction) and Qwen to
${\approx}1.2$.  Across $0.05$--$0.20$, $\rho_{\epsilon}$ moves by
under an order of magnitude.

\section{Rank-and-layer ablation sweep}
\label{app:ablation_sweep}

The causal test of \S\ref{sec:causal} extended to
$k\in\{1,3,5,10,20,50\}$.  On Llama narrow band the trajectory is
non-monotone at $k{=}5$ (Table~\ref{tab:ablation_llama}); the wide
band $[0.30L,0.85L]$ collapses to $\mathbf{0.00}$ at
$k{\geq}20$.  The per-direction decomposition
(Table~\ref{tab:perdir}) attributes the $k{=}5$ rebound to ablating
$u_1$ together with the inert $u_2$ and a weak direction, with full
collapse requiring the wider-band block---a redundancy reading
consistent with the concept cones of \citet{wollschlager2025}, though
the per-direction data alone do not exclude alternatives.

\begin{table}[h]\centering
\caption{Llama narrow-band ablation, refusal rate (Wilson $95\%$ CI),
$n_{\mathrm{gen}}{=}100$, baseline $0.97$.}
\label{tab:ablation_llama}
\begin{tabular}{@{}lccc@{}}
\toprule
$k$ & Principal & 95\% CI & Random \\
\midrule
1  & 0.85 & [0.77, 0.91] & 0.97 \\
3  & 0.80 & [0.71, 0.87] & 0.94 \\
5  & 0.92 & [0.85, 0.96] & 0.93 \\
10 & 0.66 & [0.56, 0.75] & 0.96 \\
20 & 0.10 & [0.06, 0.17] & 0.87 \\
50 & 0.03 & [0.01, 0.08] & 0.95 \\
\bottomrule
\end{tabular}
\end{table}

\paragraph{Qwen narrow-band sweep.}
On Qwen the narrow-band trajectory is monotone: refusal
$0.92\to0.54\to0.24\to0.06\to0.01\to0.01$ at $k{=}1,3,5,10,20,50$,
saturating by $k{\geq}20$---closer alignment of SVD and causal order
than Llama, but requiring more directions than Gemma ($k{=}1$).

\section{Linear-representation-hypothesis baseline}
\label{app:lrh}

$\rho_{\epsilon}$ on $M_{\mathrm{template}}$ for arbitrary
concept-difference matrices, compared to safety: alignment is only
modestly more concentrated than generic linearly represented concepts,
so the rank number is not a safety-specific signature
(Table~\ref{tab:lrh}).

\begin{table}[h]\centering
\caption{LRH baseline: $\rho_{\epsilon}$ on $M_{\mathrm{template}}$
vs.\ control concepts (mean over layers).}
\label{tab:lrh}
\begin{tabular}{@{}lcccc@{}}
\toprule
Model & EN--FR & Q--Stmt & Safety & ratio \\
\midrule
Llama-3.1-8B & $0.0078$--$0.0081$ & $0.0086$ & $0.0029$ & $2.7\times$ \\
Gemma-2-9B   & $0.0086$--$0.0104$ & $0.0090$ & $0.0048$ & $1.9\times$ \\
\bottomrule
\end{tabular}
\end{table}

\section{Calibration testbed details}
\label{app:calibration}

The Distributed variant adds to the fine-tuning loss a stable-rank
surrogate $\|M\|_F^2/\|M\|_2^2$ (a differentiable proxy for
$\rank_{\epsilon}$ that coincides with it in the flat-spectrum limit)
plus a column-orthogonality penalty, at strength $\lambda$.  Aggregate
$\rho_{\epsilon}$ on $\Ds$ (mean over $5$ seeds) is $0.008$ (Steering),
$0.173$ (Full FT), and $0.33$--$0.40$ (Distributed,
$\lambda\in\{0.5,5,15,50\}$).  Table~\ref{tab:calib_full} gives the
full projection-ablation grid behind Table~\ref{tab:lambda_sweep}
(final hidden layer, $d{=}128$, $3$ seeds).

\begin{table}[h]\centering
\caption{Post-ablation compliance on $\Ds$ after projecting top-$r$
principal directions of $M_{\Ds}$.  FullFT collapses by $r{=}80$;
Distributed $\lambda\in\{0.5,5\}$ holds $\geq0.97$ through $r{=}80$;
$\lambda{=}50$ inflates $\rho_{\epsilon}$ but is brittle.  Source:
\texttt{calibration\_v2\_revision3.json} ($3$-seed run).}
\label{tab:calib_full}
\begin{tabular}{@{}lcccccccc@{}}
\toprule
Variant & $\rho_{\epsilon}$ & $r{=}0$ & $r{=}1$ & $r{=}3$ & $r{=}10$ & $r{=}20$ & $r{=}40$ & $r{=}80$ \\
\midrule
FullFT ($\lambda{=}0$)        & $0.175$ & $0.96$ & $1.00$ & $0.84$ & $0.68$ & $0.76$ & $0.26$ & $0.07$ \\
Distributed ($\lambda{=}0.5$) & $0.332$ & $0.94$ & $0.97$ & $0.97$ & $1.00$ & $0.97$ & $1.00$ & $1.00$ \\
Distributed ($\lambda{=}5$)   & $0.400$ & $0.86$ & $0.90$ & $0.93$ & $1.00$ & $1.00$ & $1.00$ & $1.00$ \\
Distributed ($\lambda{=}15$)  & $0.404$ & $0.78$ & $0.78$ & $0.84$ & $0.88$ & $1.00$ & $1.00$ & $0.68$ \\
Distributed ($\lambda{=}50$)  & $0.405$ & $0.71$ & $0.78$ & $0.87$ & $0.93$ & $0.34$ & $0.10$ & $0.14$ \\
\bottomrule
\end{tabular}
\end{table}

\section{LLM-judge cross-check on refusal classification}
\label{app:judge}

The $15$-keyword refusal classifier (\S\ref{sec:causal}) is
cross-checked on $150$ Llama generations against a strict LLM judge:
$100\%$ agreement across baseline, narrow-band rank-$3$, and rank-$5$
conditions ($50$ each), Wilson $95\%$ CI $[0.97,1.00]$.  The $7$
judge-\texttt{COMPLIED} cases are generations in which a refusal
keyword is present but the response substantively complies with
harmful content (attack-step enumerations: Wi-Fi, brute-force,
deepfake, MITM)---the failure mode the projection ablation is designed
to expose.  Judge labels are reproducible from the released source
generations (\texttt{revision3\_llama.json}) via
\texttt{llm\_judge\_eval.py}.

\end{document}